\documentclass[10pt,journal,twocolumns] {IEEEtran}
\ifCLASSOPTIONcompsoc
\else
\fi
\ifCLASSINFOpdf
  \usepackage[pdftex]{graphicx}
  \DeclareGraphicsExtensions{.pdf,.jpeg,.png}
\else
  \usepackage[dvips]{graphicx}
  \DeclareGraphicsExtensions{.eps}
\fi
\usepackage{times}
\usepackage{helvet}
\usepackage{courier}
\usepackage{graphicx}
\usepackage{subfigure}
\usepackage{amsmath,amssymb}
\usepackage{float}

\begin{document}
%
\title{Joint Hand Detection and Rotation Estimation by Using CNN}
%
%
%
%

\author{Xiaoming~Deng, Ye~Yuan, Yinda~Zhang, Ping~Tan, Liang~Chang, Shuo~Yang, and~Hongan~Wang
\IEEEcompsocitemizethanks{\IEEEcompsocthanksitem X.M.  Deng, Y. Yuan, S. Yang and H.A. Wang are with Institute of Software, Chinese Academy of Sciences, Beijing, China
.\protect\\
E-mail: xiaoming@iscas.ac.cn, \{yuanye13,yangshuo114\}@mails.ucas.ac.cn, hongan@iscas.ac.cn
\IEEEcompsocthanksitem Y.D. Zhang is with Department of Computer Science, Princeton University, Princeton, NJ 08544, USA.\protect\\
E-mail:  yindaz@cs.princeton.edu
\IEEEcompsocthanksitem P. Tan is with School of Computing Science, Simon Fraser University, Burnaby, B.C., Canada.\protect\\
E-mail:  pingtan@sfu.ca
\IEEEcompsocthanksitem L. Chang is with College of Information Science and Technology, Beijing Normal University, Beijing, China.\protect\\
E-mail:  changliang@bnu.edu.cn
}
\thanks{Manuscript received Nov. 25, 2016.}}

\IEEEtitleabstractindextext{%
\begin{abstract}
Hand detection is essential for many hand related tasks, e.g. parsing hand pose, understanding gesture, which are extremely useful for robotics and human-computer interaction. However, hand detection in uncontrolled environments is challenging due to the flexibility of wrist joint and cluttered background. We propose a deep learning based approach which detects hands and calibrates in-plane rotation under supervision at the same time. To guarantee the recall, we propose a context aware proposal generation algorithm which significantly outperforms the selective search. We then design a convolutional neural network(CNN) which handles object rotation explicitly to jointly solve the object detection and rotation estimation tasks. Experiments show that our method achieves better results than state-of-the-art detection models on widely-used benchmarks such as Oxford and Egohands database. We further show that rotation estimation and classification can mutually benefit each other.
\end{abstract}

}

\maketitle

\IEEEdisplaynontitleabstractindextext

%
\IEEEpeerreviewmaketitle

\section{Introduction}

Hand detection is fundamental and extremely useful in human-computer interaction and robotics. It helps computers and robots to understand human intentions\cite{yang2015grasp}, and provides a variety of clues, e.g. force, pose, intention, for high level tasks. Aside from locating hands in an image, determining the in-plane rotation of the hand is also important as it is usually considered as initialization for other tasks such as hand pose estimation\cite{wang2009real} and gesture recognition\cite{wang20116d}. While generic object detection benchmarks have been refreshing over the last decade, hand detection from a single image, however, is still challenging due to the fact that hand shapes are of great appearance variation under different wrist rotations and articulations of fingers\cite{mittal2011hand}\cite{li2013pixel}.

\begin{figure*}[t]
\centering
\includegraphics[width=\textwidth]{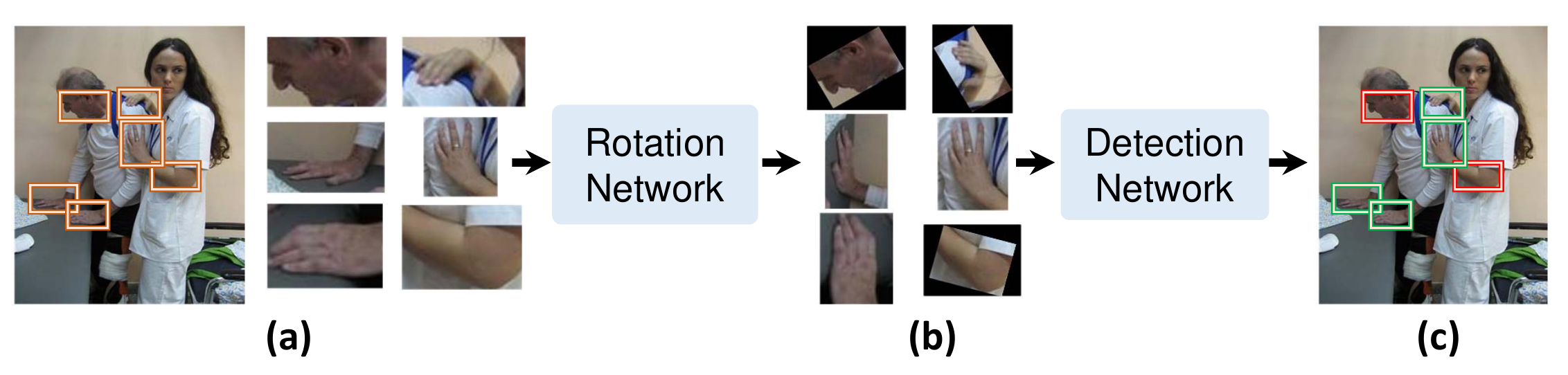}
\caption{\textbf{Pipeline of our system: Joint hand detection and rotation estimation.}
We first generate region proposals from the input image and feed them into the neural network. The in-plane rotation is estimated by the rotation network, and applied back to the input proposal. The aligned data are then fed into the detection network. Thanks to the derotation layer, two tasks are jointly optimized end-to-end.}
\label{fig0}
\end{figure*}

In this paper, we propose to solve hand detection problem jointly with in-plane rotation estimation.
Fig.~\ref{fig0} shows the general pipeline of our system.
Inspired by the RCNN \cite{girshick2014rich} framework, we first extract a number of rectangle regions that are more likely to contain a hand (Fig.~\ref{fig0}(a)). Due to the clutter of the image and the articulated shape of the hand, we propose a discriminative proposal generation algorithm, which significantly outperforms the state-of-the-art region proposal methods such as selective search\cite{uijlings2013selective} and objectness\cite{alexe2012measuring} in term of the recall.
The rotation network then estimates the in-plane rotation that align the input hand, if there is, to the upright direction (Fig.~\ref{fig0}(b)). The input data are rotated according to this estimated rotation and then fed into the detection network to perform a binary classification (Fig.~\ref{fig0}(c)).

Our model is trained jointly with multiple tasks simultaneously, which has been demonstrated to be very successful for many vision tasks. In our case, hand detection and in-plane rotation are closely related and could benefit each other. Calibrating training data under different rotation to upright position results in rotation invariant feature, which relieves the burden of the detection/classification model. While in return, detection model can verify if the rotation estimation is reasonable. However, due to the nature of the convolutional neural networks, rotation invariance is more difficult to achieve than translation invariance, which prevents us from an end-to-end optimization. As a result, previous works \cite{rowley1998rotation}  usually handle transformation estimation and detection separately or in a iterative fashion, which may not achieve a global optima.

We design a derotation layer, which explicitly rotates a feature map up to a given angle. This allows us to jointly optimize the network for two tasks simultaneously (See Fig.~\ref{fig1} for the network structure). Recently, spatial transformer networks (ST-CNN) \cite{jaderberg2015spatial} also presented a differentiable module to actively spatially transform feature maps with CNN. However, their transformation is learned unsupervised such that could be any arbitrary rotation that are not directly interpretable(The discussion that ST-CNN may not be the ideal hand detection model are shown in the appendix). Also, the transformation space is typically huge and would require much more data and time to converge. Comparatively, our rotation estimation network is aimed for upright alignment, such that the output can be directly used for related tasks, e.g. hand pose estimation. It is also trained supervised, which is more likely to converge.

The contributions of this paper are mainly in four aspects. First, we propose, by our knowledge, the first framework that jointly estimates the in-plane hand rotation and performs detection. Experiment shows that we achieve significant better performance than state-of-the-art on widely used benchmark. Second, we design the derotation layer, which allows end-to-end optimization with two tasks simultaneously. The rotation estimation network is trained with strong supervision, which converges more efficiently. Third, we propose a hand shape proposal generation algorithm with significantly improved recalls and Mean Average Best Overlap score(MABO) outperforming the state-of-the-art selective search\cite{uijlings2013selective} and objectness\cite{alexe2012measuring}. Last but not least, our model is much more efficient than previous work. Thanks to the rotation estimation network, we do not need to brute force search for all possible angles and thus reduce the detection time to $1/7$.

\section{Related Work}
Recent hand detection methods from a single image can be classified into four categories:

\noindent \textbf{Skin Detection Method.} These methods build a skin model with either Gaussian mixture models \cite{sigal2004skin}, or using prior knowledge of skin color from face detection\cite{viola2001robust}. However, these methods often fail to apply to hand detection in general conditions due to the fact that complex illuminations often lead to large variations in skin color and make the skin color modelling problem challenging.

\noindent \textbf{Template Based Detection Method.} These methods usually learn a hand template or a mixture of deformable part models. They can be implemented by Harr features like Viola and Jones cascade detectors \cite{dalal2005histograms}, HOG-SVM pipeline\cite{dalal2005histograms}, mixtures of deformable part models(DPM) \cite{mittal2011hand}. A limitation of these methods is their use of weak features (usually HOG or Harr features). There are also methods that detects human hand as a part of human structure, which uses the human pictorial structure as spatial context for hand position \cite{karlinsky2010chains}. However, these methods require most parts of human are visible, and occlusion of body parts makes hand detection difficult \cite{ghiasi2014parsing}.

\noindent \textbf{Per-pixel Labeling Detection Method.} A pixel labeling approach \cite{li2013pixel} has shown to be quite successful in hand detection in ego-centric videos. In \cite{zhu2014pixel}, the pixel labeling approach is further extended to a structured image labeling problem. However, these methods require time-consuming pixel-by-pixel scanning for whole image.

\noindent \textbf{Detection Method with Pose Estimation.} These methods can be classified as two types: 1) first estimate the object pose, and then predict the object label of the image derotated with the object pose;
Rowley, Baluja, and Kanade\cite{rowley1998rotation} proposed a seminal rotation invariant neural network-based face detection. The system employs multiple networks: the first is a rotation network which processes each input window to determine its orientation, and then uses this information to prepare the window for one or more detector networks. 2) simultaneous pose estimation and detection. He, Sigal and Sclaroff \cite{he2014parameterizing} proposed a structured formulation to jointly perform object detection and pose estimation. Fidler et. al.\cite{fidler20123d} proposed a 3D object detection and viewpoint estimation with a deformable 3D cuboid model. As far as we know, less attention is paid on using convolutional neural networks to jointly model object detection and rotation estimation problems for 2D images.

\section{Approach}
\begin{figure*}[t]
\centering
\includegraphics[width=\textwidth]{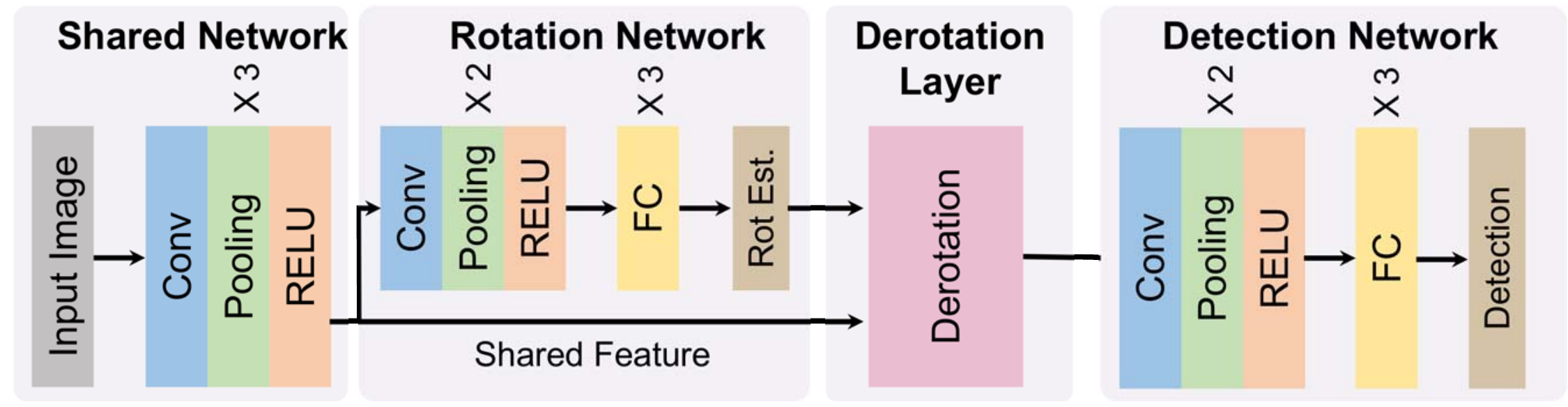}
\caption{\textbf{Overview of our model.} The network consists of four parts: 1) a shared network for learning features to benefit both rotation estimation and detection tasks; 2) a rotation network for estimating the rotation of a region proposal; 3) a derotation layer for rotating inputted feature maps to a canonical pose; 4) a detection network for classifying the derotated proposal. These modules are jointly used to learn an end-to-end model for simultaneous hand detection and rotation estimation.}
\label{fig1}
\end{figure*}

We present an end-to-end optimized deep learning framework for joint hand detection and rotation estimation with a single image.
The overall pipeline is illustrated in Fig.~\ref{fig1}.
We first extract proposals for regions that are likely to contain a hand. To particularly take the advantage of the strong local characteristic of hand shape and color, we train a multi-component SVM using features from Alexnet\cite{krizhevsky2012imagenet} for region proposal rather than simple segmentation based algorithm.
The proposals are fed into convolution layers to exact shared features that will be used for both rotation estimation and detection afterward.
The rotation network performs a classification task to estimate an in-plane rotation that could align the hand in the input image, if any, to the upward direction. Then, the shared feature is explicitly rotated according to the result from the rotation network, and pass through the detection network for a confidence score.
Since the feature is supposed to be well aligned, the detection network does not need to handle the alignment and thus performs more reliably.
The rotation transformation is done by the derotation layer, which allows back propagation and enable an end-to-end training. Different to ST-CNN\cite{jaderberg2015spatial}, both the rotation network and detection network are trained under supervision, therefore the output of the rotation network is guaranteed for the desired data alignment.


\subsection{Proposal Generation}
A variety of category-independent region proposals are proposed including selective search\cite{uijlings2013selective}, objectness\cite{alexe2012measuring}, and category independent object proposals\cite{endres2010category}.
However, due to the cluttered background and articulated shape of the hands, previous region proposals (especially segmentation based) can no longer guarantee recall while keep the number of proposals to be manageable(Table 1 from experiment section shows that selective search and objectness are both not good at hand detection.).

We adopt a discriminative approach to generate hand region proposal. Inspired by deformable parts model(DPM)\cite{felzenszwalb2010object}, we cluster the training data to 8 subgroups based on the aspect ratio of the image patches, and train one linear SVM using pooled conv5 layer feature extracted from the Alexnet \cite{krizhevsky2012imagenet} for each group. We learn the threshold on validation set such that 100 percent of the positive data is covered with at least 0.5 Intersection over Union(IOU). Fig.\ref{figproposala} shows that our method ensures significantly higher recall while keeps the number of proposal per image comparable.


\subsection{Rotation Aware Network}
In this section, we introduce the rotation aware neural network to decide if a region proposal contains a hand. The detailed network structure is in supplementary material.
\subsubsection{Network Structure}
The network starts with 3 convolution+relu+pooling to extract features from input proposal patches. The resolution of the input feature map is $13\times13$  due to the strides but still maintains the spatial information. Built upon this feature, the rotation network consists of 2 convolution layers followed by 3 fully connected layer and estimate the angles to rotate such that the hand, if any, in the proposal could be aligned to the upward direction.
We formulate the rotation estimation problem into a regression problem. Given an rotated hand, the rotation network performs as a regressor and outputs a two dimensional rotation estimation vector $\mathbf{l} = (\cos\alpha,\sin\alpha)$. The estimated $\mathbf{l}$ will then be sent to the derotation layer to rectify the orientation of training patches. Afterward, a derotation layer rotates the feature from the shared network according to the estimated in-plane angle from the rotation network. The rotated feature is then fed into 2 convolution layers and 3 fully connected layers to perform a binary classification, telling if the proposal contains a hand. Since the derotation layer is differentiable, the whole network can be optimized end-to-end, and two tasks, rotation estimation and hand detection, can be jointly optimized.

\subsubsection{Derotation Layer}
\begin{figure}[t]
\centering
\includegraphics[width=0.4\textwidth]{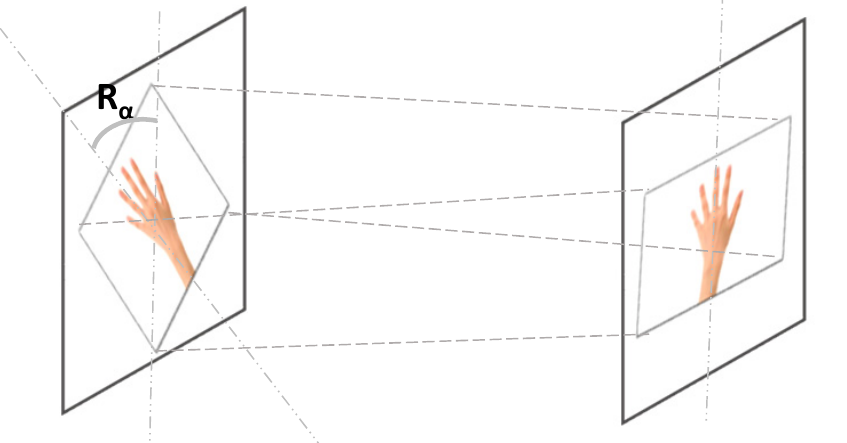}
\caption{Illustration of applying derotation transformation to input feature map. The derotation layer aims to warp the input image to a canonical hand pose by $\mathbf{R}_\alpha$. In this work, the canonical hand pose is an upright hand pose as shown in the right part of this figure.}
\label{fig3}
\end{figure}

\begin{figure}[h]
 \centering
\begin{minipage}{0.49\linewidth}
\begin{center}
\subfigure[]{\includegraphics[width=\textwidth]{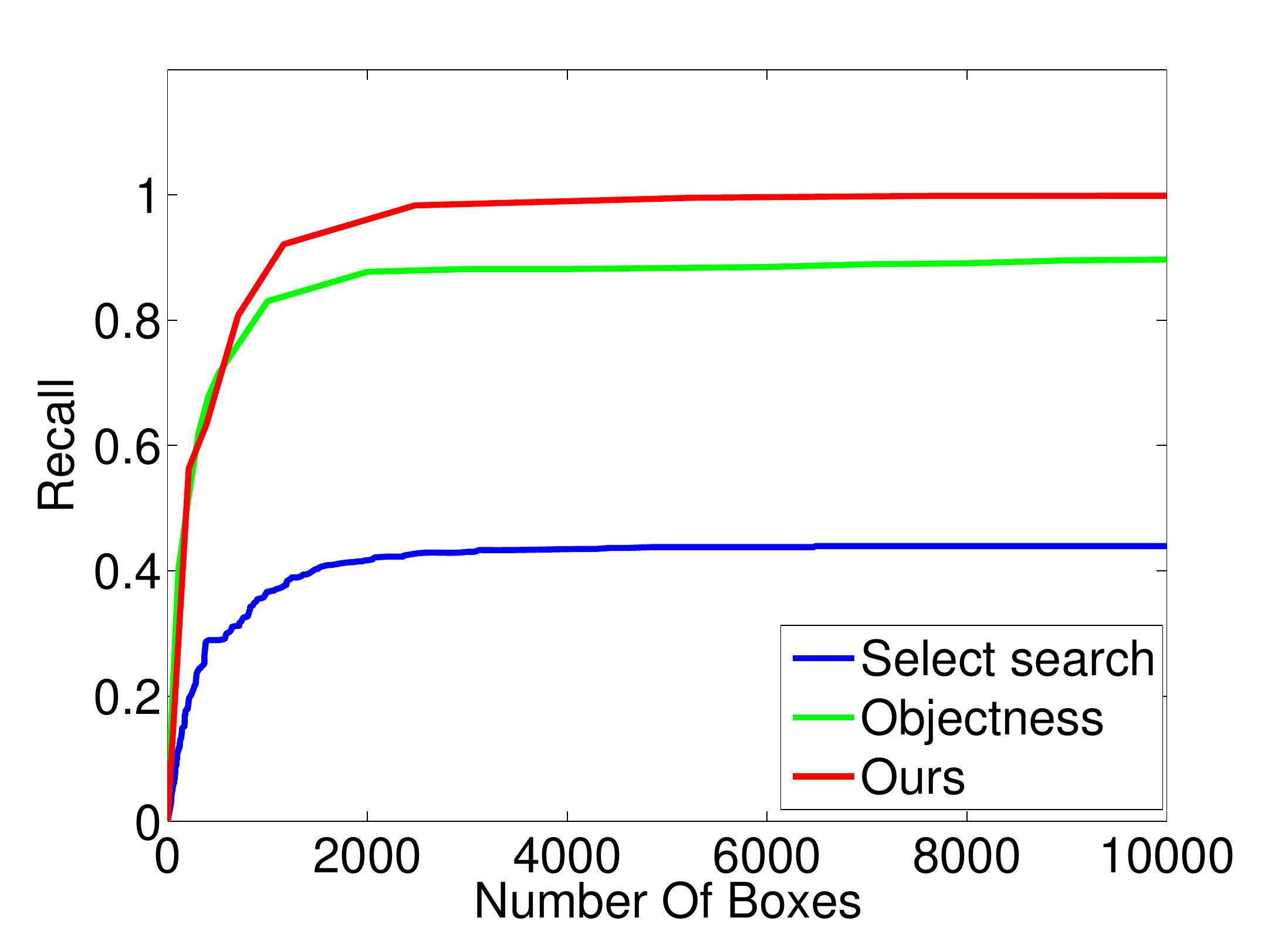}\label{}}
\end{center}
\end{minipage}
\begin{minipage}{0.49\linewidth}
\begin{center}
\subfigure[]{\includegraphics[width=\textwidth]{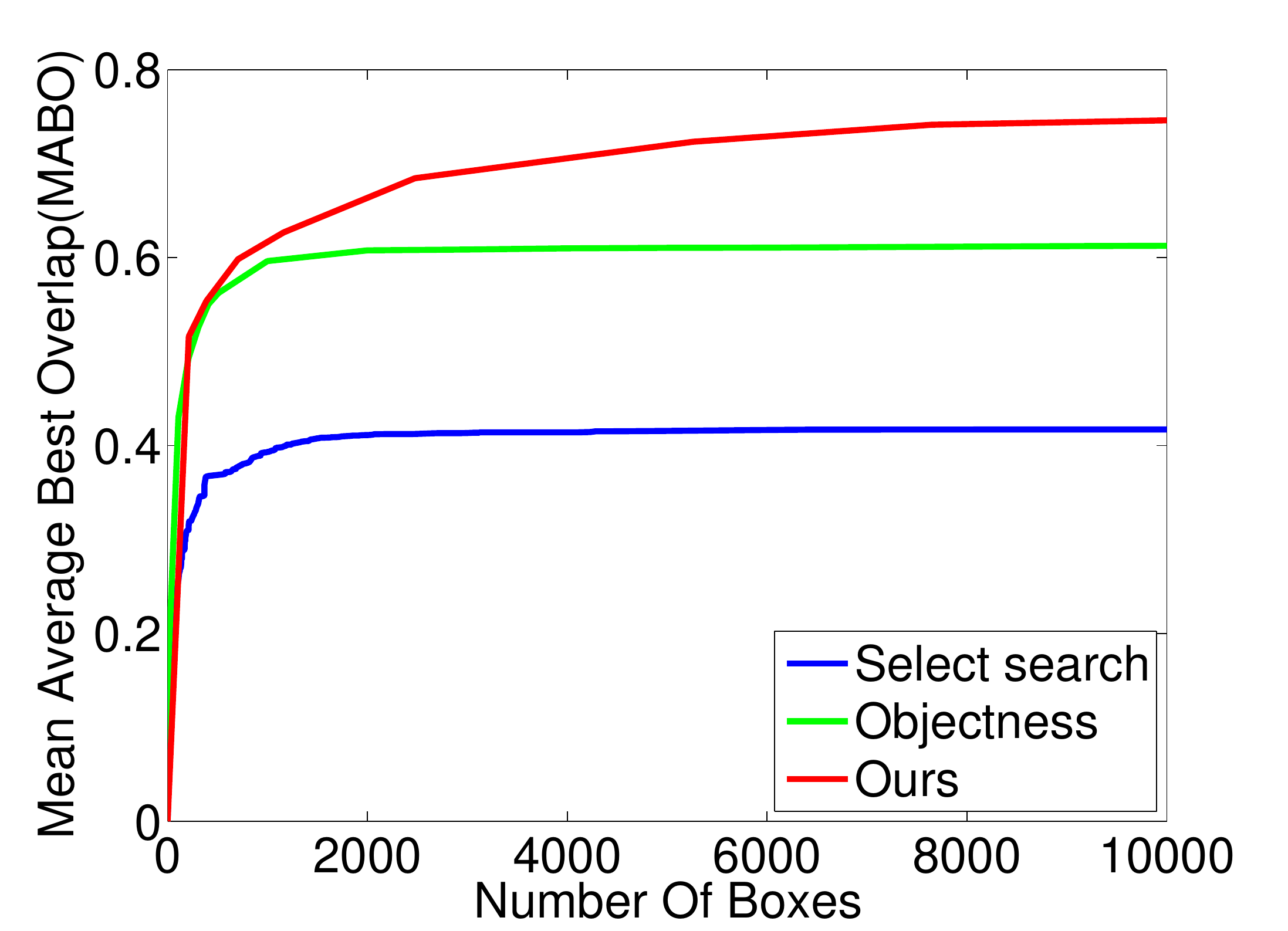}\label{}}
\end{center}
\end{minipage}
\caption{Trade-off between recall(a) and MABO(b) of the object hypotheses in terms of bounding boxes on the Oxford hand test set.}
\label{figproposala}
\end{figure}

Derotation layer is a layer which applies a rotation transformation to a feature map during a single forward pass.
In our scenario, the input of a derotation layer is the feature map computed from the original image and a in-plane rotation angle predicted from either the rotation network or ground truth, and the output of this layer is the derotated feature map under the given rotation angle, while supposedly under the canonical upright hand pose (Refer to Fig.~\ref{fig3}).

Specifically, if $\alpha$ is the in-plane rotation angle we want to apply, the derotation transformation is
\begin{equation}
\label{eq4}
\left[ {{\begin{array}{*{20}c}
 x' \\
 y'
\end{array} }} \right] = \underbrace{\left[ {{\begin{array}{*{20}c}
 \cos\alpha & -\sin\alpha\\
 \sin\alpha & \cos\alpha
\end{array} }} \right]}_{\mathbf{R}_\alpha}\left[ {{\begin{array}{*{20}c}
 x \\
 y
\end{array} }} \right]
\end{equation}
\noindent where $[x',y']$ is the target coordinates of the regular grid in the output feature map under the canonical upright hand pose, $[x,y]$ is the source coordinates of the regular grid in the input feature map.

In our implementations, we use inverse mapping to get the output feature map. In other word, for each pixel $[x',y']$ of the output we get the corresponding position $[x, y]$ in the input feature map. Since $[x, y]$ is often not located on a regular grid, we calculate the feature by averaging the values from its four nearest neighbor locations. We pad zero to $[x, y]$, which is outside of the regular grid. An alternative choice is to pad with feature outside the box, which is not implemented here due to efficiency issue.

The back-propagation can be done with a record of the mapping between coordinates between feature map before and after the derotation layer. Updating value on $[x', y']$ is backward propagated to the coordinates from which its value comes, which is in a similar fashion as some pooling layer and ROI layer in \cite{girshick2015fast}\cite{ren2015faster}.

\subsection{Loss Layer}
Our model overall has two losses, the rotation network loss and the detection network loss.

\noindent \textbf{Rotation loss.} For rotation estimation, we use $L_2$ loss. We get ground true hand bounding boxes and use them to train a network that can
do regression on the hand's rotation, formulated as a two-dimensional vector $\mathbf{l} = (\cos\alpha,\sin\alpha) \triangleq (\frac{c}{\sqrt{c^2 + s^2}},\frac{s}{\sqrt{c^2 + s^2}}) $. Here, $c,s$ are outputs of the final convolutional layer in rotation network, $\mathbf{l}$ is enforced as a normalized pose vector by normalizing $c,s$, and thus we can enforce $\mathbf{R}_\alpha$ in Eq.(\ref{eq4}) as a rotation matrix. More exactingly, if $\mathbf{l}$
and $\mathbf{l}^{*}$ are the predicted and ground truth rotation estimation vectors, the rotation loss is
\begin{equation}
\label{eq5}
L_{rotation}(\mathbf{l},\mathbf{l}^{*}) = ||\mathbf{l} - \mathbf{l}^{*}||_2^2
\end{equation}

It is easy to compute the partial derivative of $L_{rotation}$ w.r.t $\mathbf{l} = (\cos\alpha$, $\sin\alpha)$. To deduce the backward algorithm of rotation loss, we need to compute the partial derivative of $\mathbf{l} = (\cos\alpha,\sin\alpha)$ w.r.t. $c,s$, which can be calculated as follows:
\begin{eqnarray}
\nonumber \frac{\partial \cos\alpha}{\partial c} &=& (c^2 + s^2)^{-\frac{1}{2}} - c^2(c^2 + s^2)^{-\frac{3}{2}}\\
\nonumber \frac{\partial \cos\alpha}{\partial s} &=& -cs(c^2 + s^2)^{-\frac{3}{2}}\\
\nonumber \frac{\partial \sin\alpha}{\partial c} &=& -cs(c^2 + s^2)^{-\frac{3}{2}}\\
\nonumber \frac{\partial \sin\alpha}{\partial s} &=& (c^2 + s^2)^{-\frac{1}{2}} - s^2(c^2 + s^2)^{-\frac{3}{2}}
\end{eqnarray}

\noindent \textbf{Detection loss.} For detection task, we use a simple cross-entropy loss. Denote $D^{*}$  to be the ground truth object labels, and we use
the detection loss as follows
\begin{equation}
\label{eq6}
L_{detection}({D},{D}^{*}) = - \frac{1}{n} \sum_i \sum_j D_i^{*} \log(D_i)
\end{equation}
\noindent where $D_i = {e^{z_j^i}}/{\sum_{j = 0}^1e^{z_j^i}}$ is the prediction of class $j$ for proposal $i$ given the output $z$ of the final convolutional layer in detection network, $n$ is the training proposal number.

\subsection{Training Schema}
The rotation aware network contains two pathways that interact with each other through the derotation layer. To train the model, we adopt a divide and conquer strategy. We first initialize the shared network and the rotation network with the model pretrained on ImageNet, and only fine tune on the rotation estimation task. Then, we fix these two networks but enable the detection network, and fine tune for the hand binary classification task. After the network parameters converge to reasonable good local optima, we enable all the network and optimize in a end-to-end manner for both tasks.

We take only the region proposals from proposal generation section as the training data. Depending on the IOU with ground truth, a region proposal is considered to be positive if the IOU is larger than 0.5; negative if the IOU is smaller than 0.5; discarded otherwise, which ends up with about 10K positives and 49 million negatives. Since the number of positive and negative data is extremely imbalanced, we use all the positives and randomly sampled 30 million negatives. We also ensure the ratio between positive and negative data in each mini-batch to be 1:1. For the negative data, they do not contribute any gradient during the back propagation from the rotation network.

\subsubsection{Data Augmentation}
Since the pretrained Alexnet on ImageNet does not encode rotation related information, we found that training with the limited number of positive data results in poor generalization capability. We relieve the overfitting by augmenting the size of the training data. We horizontally flip the training image, which allows invariance against left/right hand. We also rotate each positive proposals to 36 times (10 degree once) around the center of the patch, and take each of them as a training data. This dramatically increases the size of training data to 36 times bigger, and our training data eventually contains over 6 million proposals from 4096 images.

\subsubsection{Hard Negative Mining}
Inspired by DPM\cite{felzenszwalb2010object}, we perform hard negative mining to suppress confusing false alarms. After training a model from the initial training data, we run the model on the whole training proposals. We add the negatives with high score ($\geq0.4$) to the training data and train a model again. Generally speaking, 2-3 times of hard negative mining are sufficient to improve the performance to be almost saturated.

\section{Experiments}

\subsection{Dataset and Evaluation}
The proposed method is evaluated on widely-used Oxford hand dataset\cite{mittal2011hand} and EgoHands dataset\cite{egohands2015iccv}. The Oxford hand dataset contains 13050 hands annotated with bounding box and rotation from images collected from various public image datasets. The dataset is considered to be diverse as there is no restriction imposed on the pose or visibility of people, and background environment. Oxford hand dataset has much cluttered background, more viewpoint variations and articulated shape changes than other popular hand dataset such as Signer\cite{Signer} and VIVA\cite{VIVA}. The EgoHands dataset \cite{egohands2015iccv} contains 48 Google Glass videos of complex, first-person interactions between two people, which are also annotated with bounding box and rotation. This dataset is mainly used to recognize hand activities in first-person computer vision.

To demonstrate the performance of the whole system, we evaluate the region proposal generation, the rotation estimation, and final detection performance respectively.
For region proposal generation, we measure the percentage of positive data that is covered by any proposal with an IOU larger than 0.5. To further show the localization precision, we also calculate the Mean Average Best Overlap (MABO)\cite{uijlings2013selective}, which is a standard metric to measure the quantity of the object hypotheses. For rotation estimation, we measure the difference between the estimation and the ground truth. For the detection, we use the typical average precision(AP) and precision-recall curve with a threshold 0.5 on IOU.


\subsection{Performance on Oxford Hand Dataset}
\subsubsection{Region Proposal Generation}
\begin{figure}[b]
 \centering
 \begin{tabular}{cc}
\includegraphics[width=0.6\linewidth]{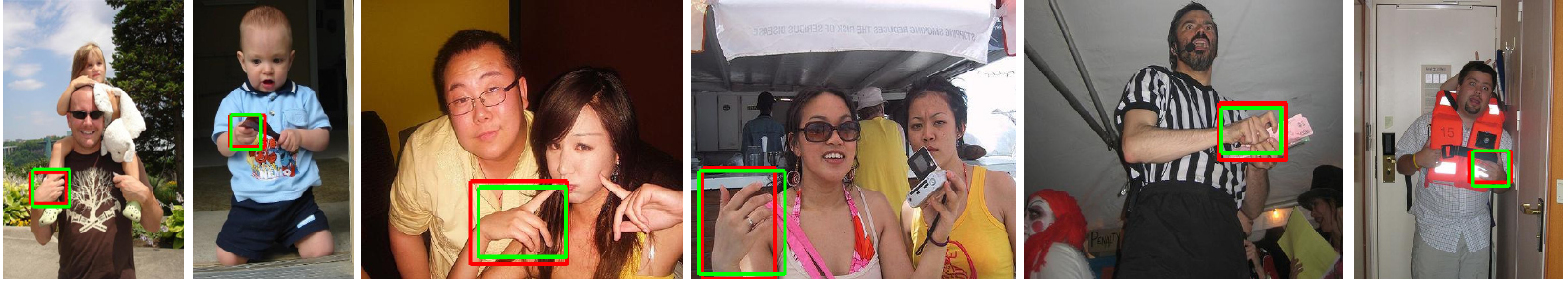} & Ours \\
\includegraphics[width=0.6\linewidth]{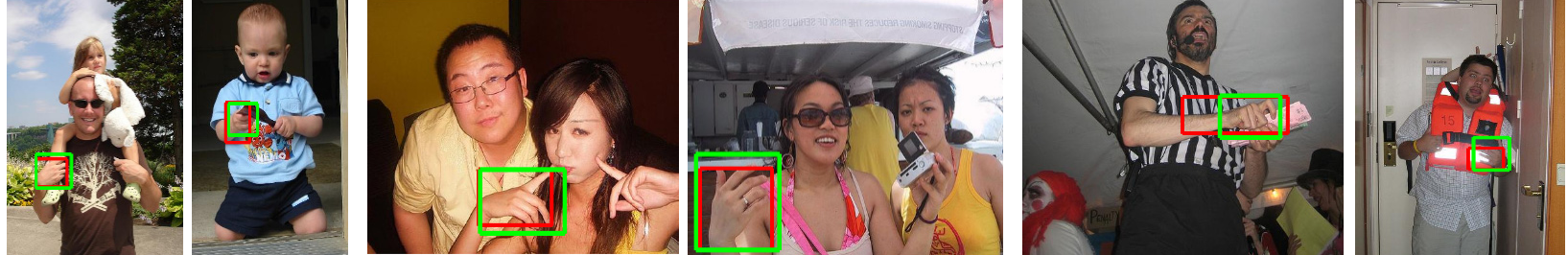} & Objectness\\
\includegraphics[width=0.6\linewidth]{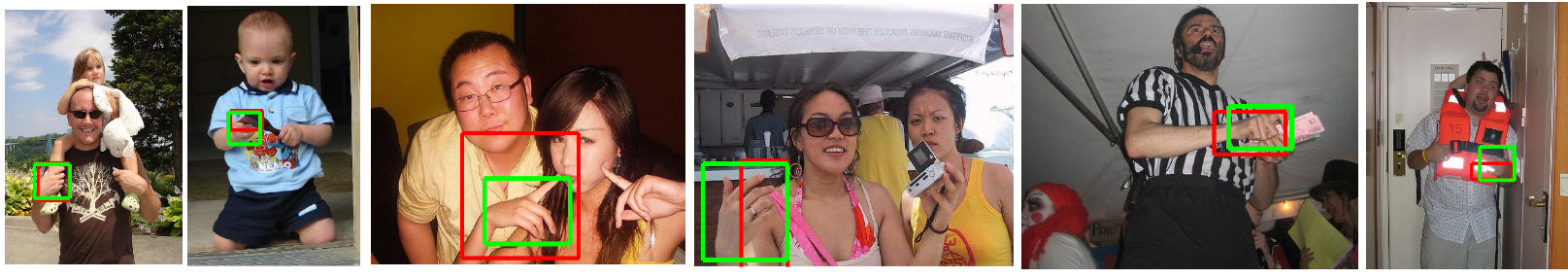} &Selective search
\end{tabular}
\caption{Comparison between our proposal generation approach to the traditional segmentation based algorithms. Examples of locations for objects whose Best Overlap score is around MABO of our method, objectness and selective search. The green boxes are the ground truth. The red boxes are created using our proposal generation method.}
\label{figproposalb}
\end{figure}

Fig. \ref{figproposalb} and Table \ref{tab2} show comparisons between our SVM based approach to the traditional segmentation based algorithms such as selective search\cite{uijlings2013selective} and objectness\cite{alexe2012measuring}. We achieve nearly 100\% recall and a significantly higher MABO with only about half the number of proposals (7644 vs. 13000+) used in selective search and objectness. Qualitatively, selective search fails due to the fact that it relies much on over-segmentation and may not be suitable for complex scenarios with cluttered background and many connected skin-like regions, while our method could take advantage of the discriminative power of the articulated local shape of the hand and generate reliable proposals.

\begin{table}[htbp]
\caption{Proposal generation performance on the hand test data. \#win means the window numbers.} \centering
\begin{tabular}{c|ccc}
  \hline
  {Method} & Recall & MABO &  \#win\\
  \hline
  selective search &  46.1\% &  41.9\% & 13771 \\
  objectness &  90.2\% &  61.6\% & 13000 \\
  our proposal method &  99.9\% &  74.1\% & 7644\\
  our proposal method &  100\% &  76.1\% & 17489\\
  \hline
\end{tabular}
\label{tab2}
\end{table}


\subsubsection{Rotation Estimation}
\begin{figure*}[t]
\centering
\includegraphics[width=0.75\textwidth]{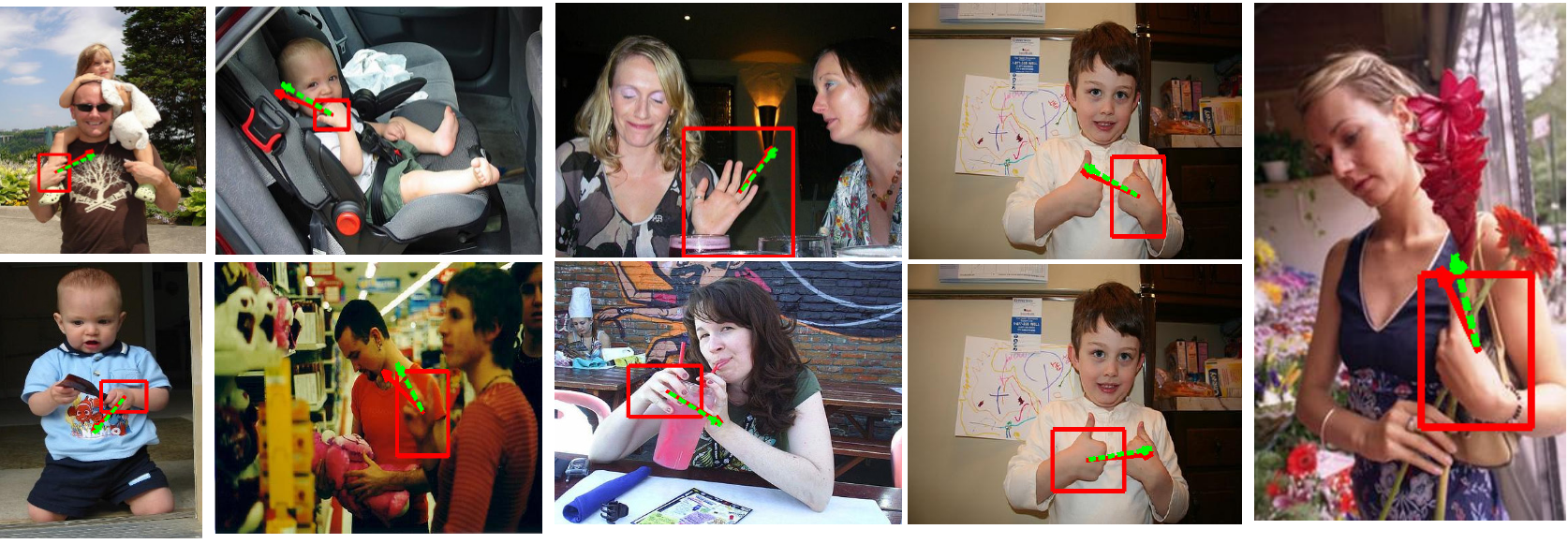}
\caption{Examples of hand rotation estimation results on proposals of test images. The red and green arrows indicate the estimated and ground truth rotation angles, respectively.}
\label{fig:rotation}
\end{figure*}

We first demonstrate that the rotation network can produce reasonable in-plane rotation estimation. Table \ref{tablerotation} shows the performance of the rotation estimation(Refer to only rotation model). We can see that the prediction for 45.61\% of the data falls in 10 degree around the ground truth, and 70.13\% for 20 degree, 79.79\% for 30 degree. Examples of hand rotation estimation results on test images are also shown in Fig. \ref{fig:rotation}. We see that our rotation model leads to excellent performance.

\begin{table}
\caption{Rotation estimation performance on the hand test data. Rotation is correct if the distance in degree between prediction and ground truth is less than $\delta_\alpha = 10^\circ, 20^\circ, 30^\circ$. We compare the rotation estimation results on the hand test data with only rotation model, and joint rotation and detection model. } \centering
\begin{tabular}{c|ccc}
  \hline
  {Method} & $\leq 10^\circ$ & $\leq 20^\circ$ & $\leq 30^\circ$ \\
  \hline
  Only rotation model &  45.61\% & 70.13\% & 79.79\%\\
  Joint model    &  47.84\% & 70.88\% & 80.24\%\\
  \hline
\end{tabular}
\label{tablerotation}
\end{table}

\subsubsection{Detection Performance}
We compare our model to several state-of-the-art approaches such as R-CNN\cite{girshick2014rich}, DPM-based method\cite{mittal2011hand}, DP-DPM\cite{girshick2015deformable} and ST-CNN\cite{jaderberg2015spatial}, the first three of which do not explicitly handle rotation. Fig. \ref{precision_recall}(a) shows the precision recall curves, and the number after each algorithm is the average precision(AP).
Our model(seperated) means that the shared 3 convolution layers are kept unchanged, and the rotation and detection networks are trained separately with shared network not tuned, and our model(joint) means that the network is end-to-end trained. Our average precision on Oxford hand dataset is 48.3\% for our model(joint), which is significantly better (11.5\%, 6\% higher) than the state of the art\cite{mittal2011hand}, in which AP = 36.8\% is reached with DPM trained with hand region, and AP = 42.3\% is reached with additional data such as hand context and skin color model(We do not use such additional data). Our models, joint or separated, is advantageous over seminal CNN-based methods, AP of our separated model is 4.9\% higher than R-CNN, 6.6\% higher than ST-CNN. This demonstrates that data alignment with rotation is very critical for the classification model in the detection network. In Fig. \ref{figresults}, we show some results of our method on test image from Oxford hand dataset, in which both detection bounding boxes and rotation estimation results are shown. The discussion that ST-CNN may not be the ideal hand detection model is shown in the appendix.

\begin{figure*}[htbp]
 \centering
 \begin{tabular}{cc}
\includegraphics[width=0.45\linewidth]{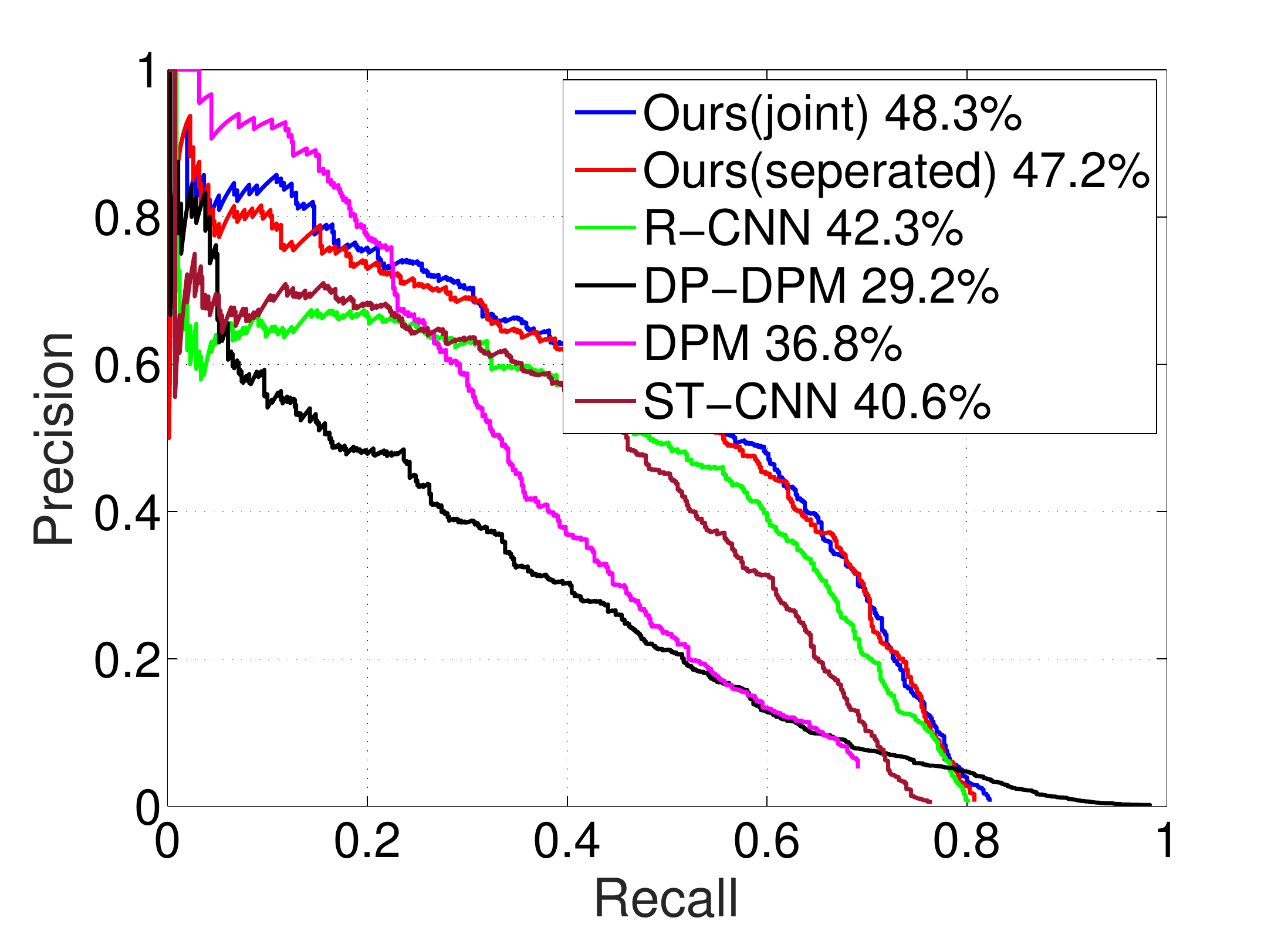} & \includegraphics[width=0.45\linewidth]{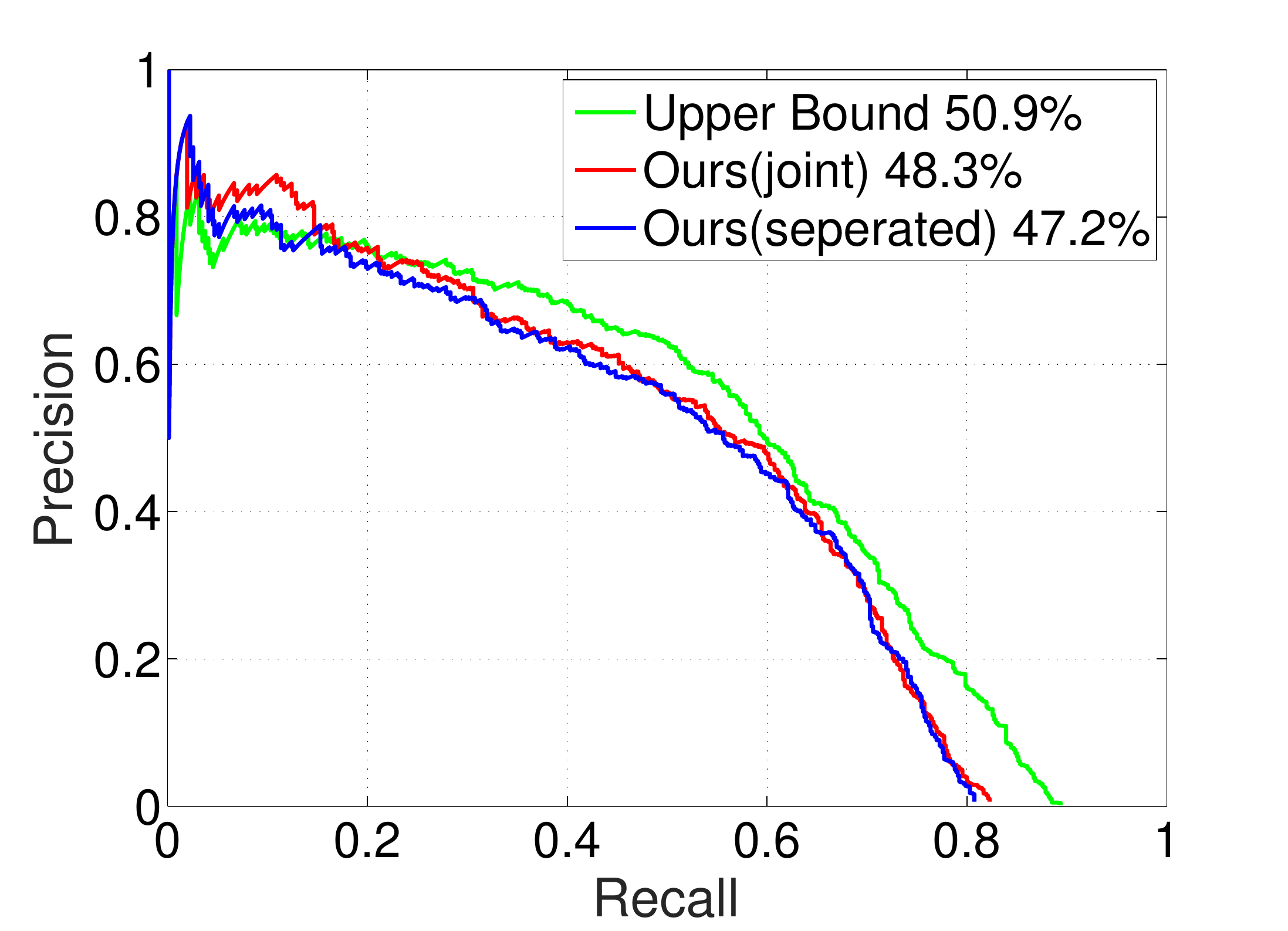} \\
 (a) & (b)
\end{tabular}
\caption{Precision-Recall curve comparing the baseline and final results. (a) Comparison with baselines. DPM means the results with hand shape detector in \cite{mittal2011hand}. (b) Comparison with detection with ground truth rotation, a performance upper bound.}
\label{fig:precision_recall}
\end{figure*}

\begin{figure*}[htbp]
\centering
\includegraphics[width=0.9\textwidth]{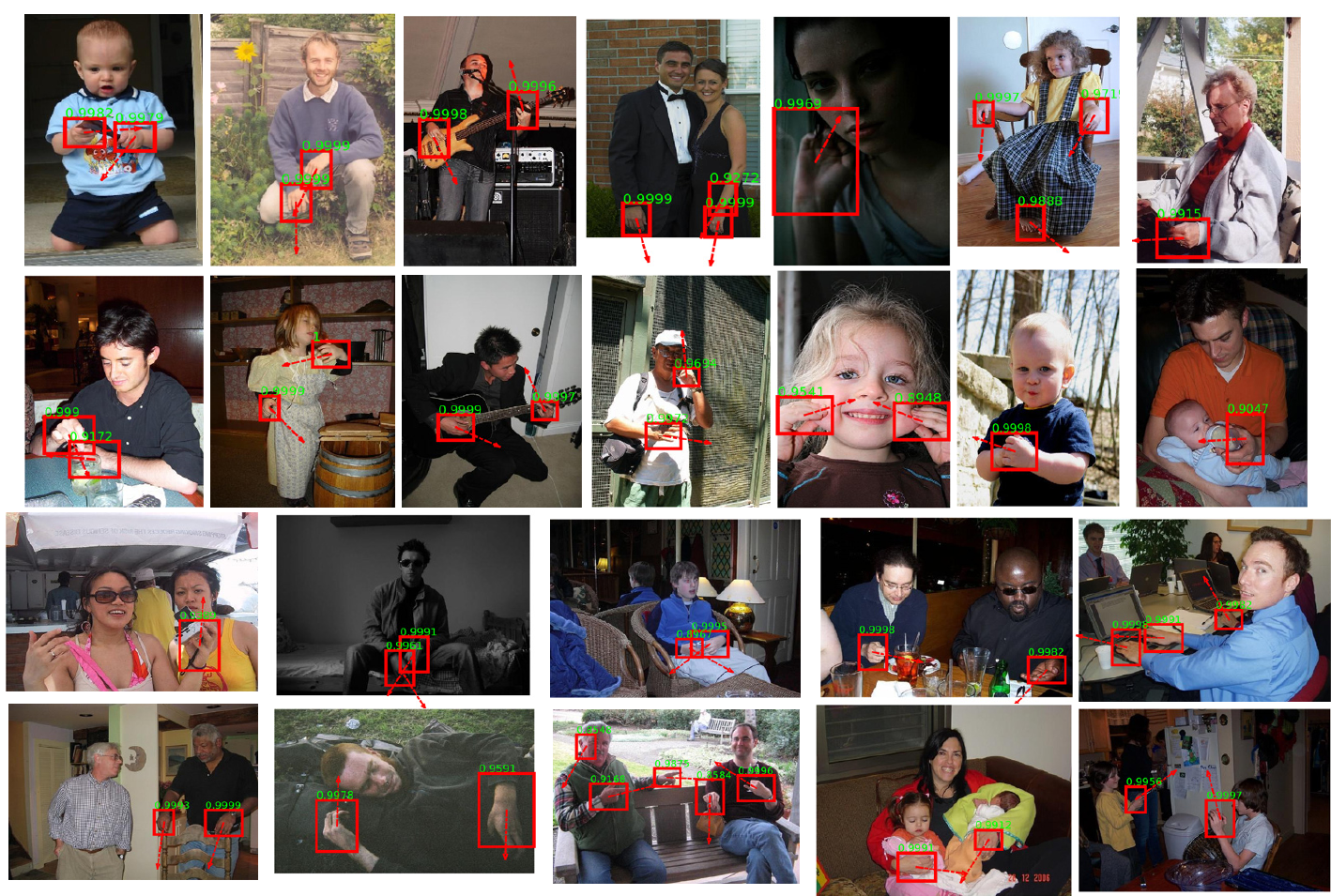}
\caption{Examples of high-scoring detection on Oxford hand database. The rotation estimation is illustrated with red arrows.}
\label{figresults}
\end{figure*}

\begin{figure*}[htbp]
\centering
\includegraphics[width=0.5\textwidth]{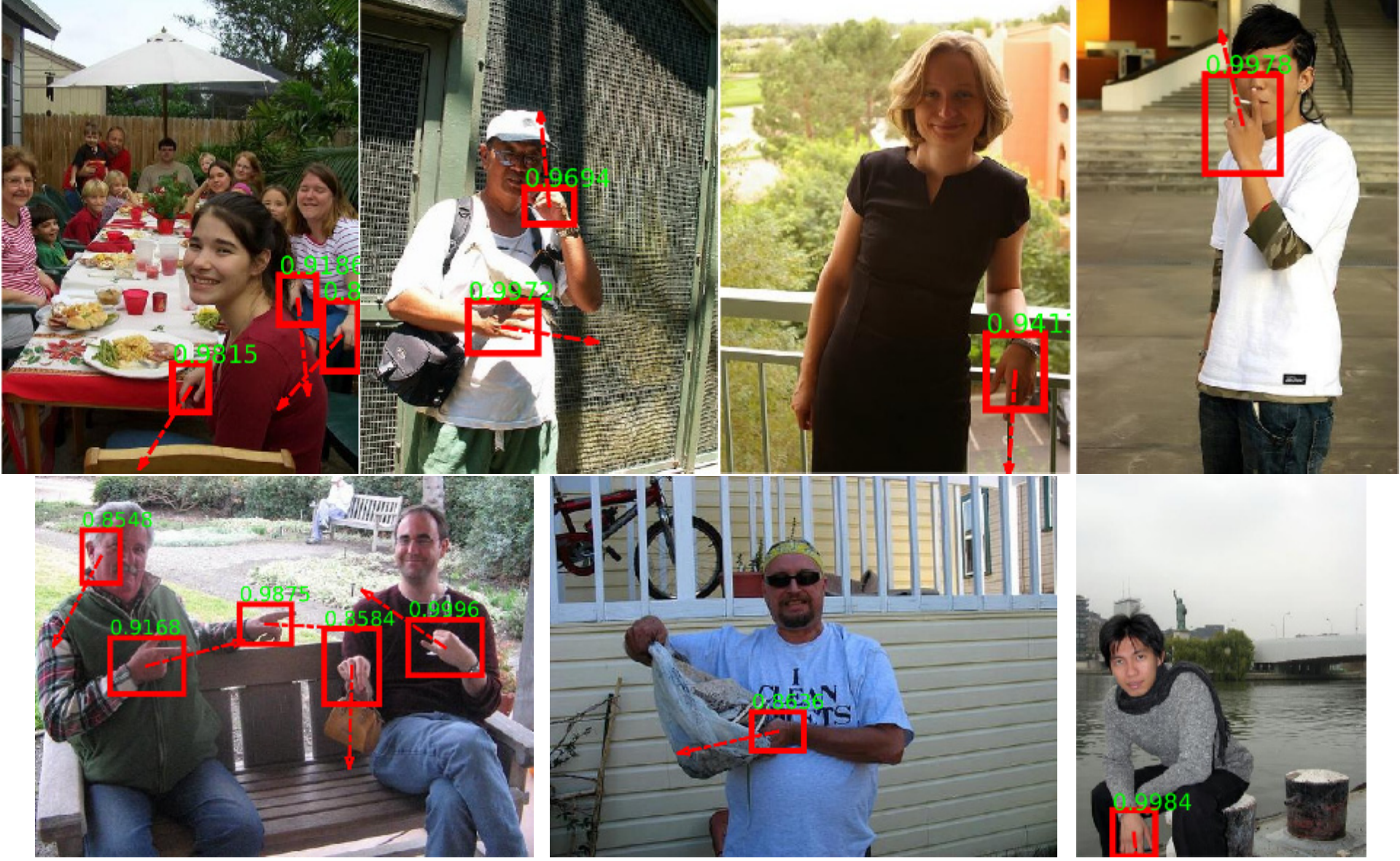}
\caption{Examples of high-scoring detection on Oxford hand database(outdoor images). The rotation estimation is illustrated with red arrows.}
\label{figoxfordoutdoor}
\end{figure*}

\begin{figure*}[t]
\centering
\includegraphics[width=0.8\textwidth]{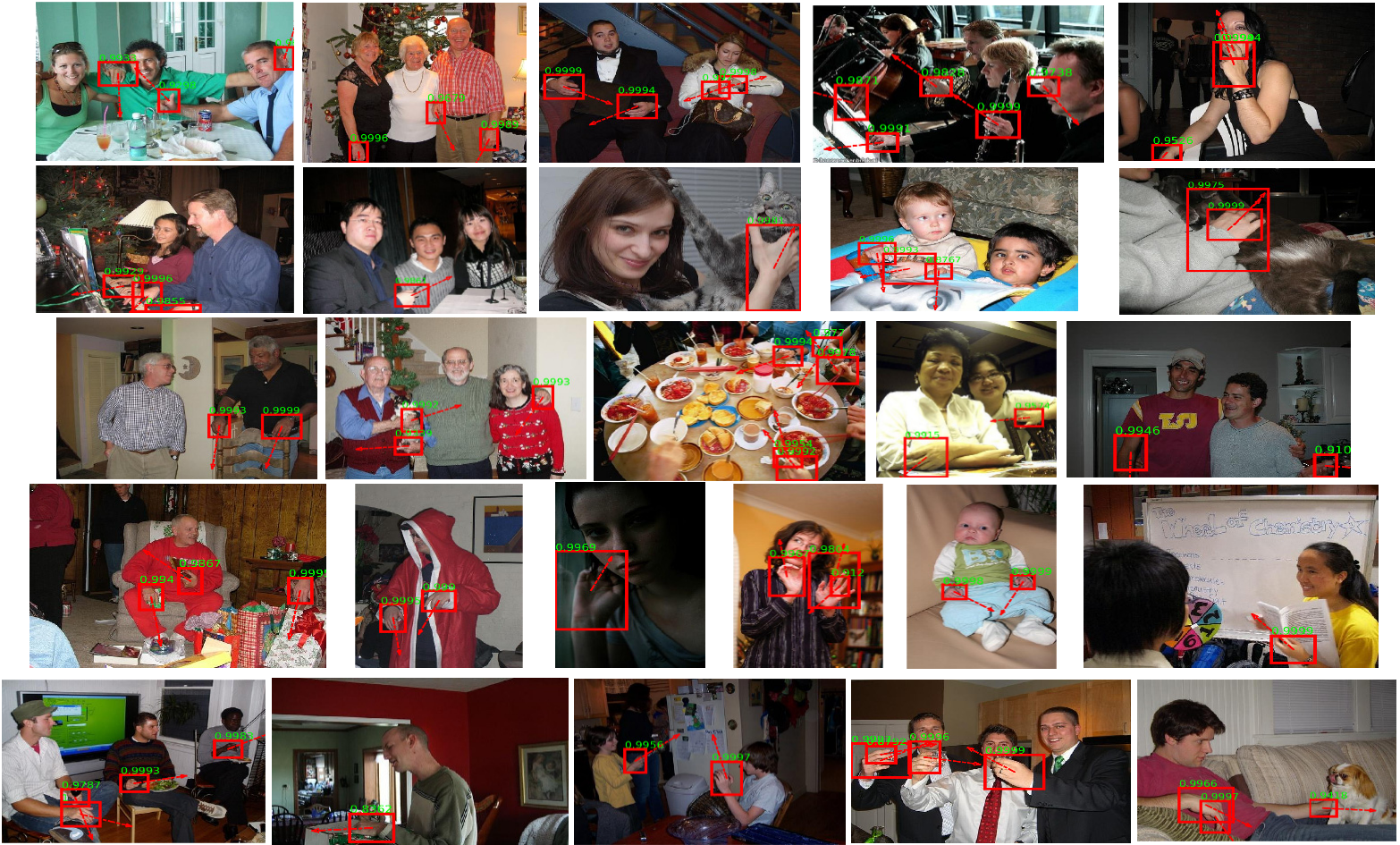}
\caption{Examples of high-scoring detection on Oxford hand database(indoor images). The rotation estimation is illustrated with red arrows.}
\label{figoxfordindoor}
\end{figure*}

\begin{figure*}[htbp]
\centering
\includegraphics[width=0.8\textwidth]{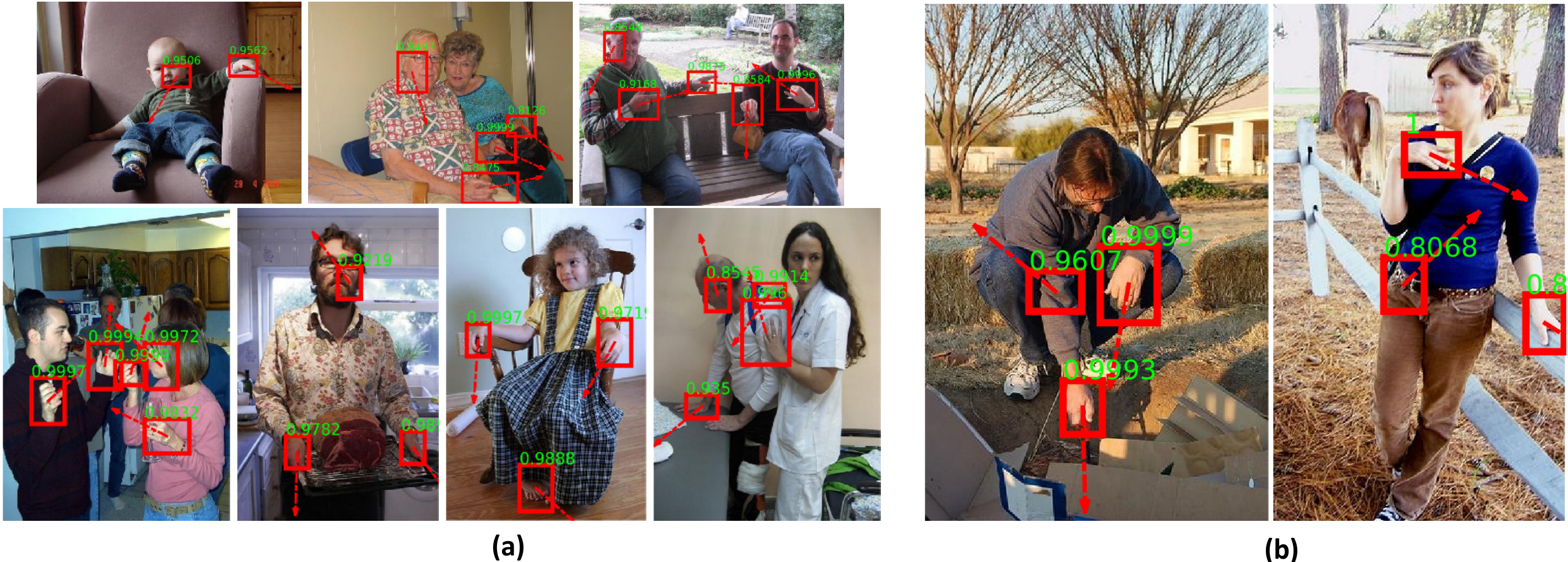}
\caption{Examples of false alarm detection on Oxford hand database. (a) false alarm with skin-like region. (b) false alarm with non-skin region.}
\label{figoxfordfalse}
\end{figure*}

We give more experimental results of our hand detector on Oxford hand dataset. Fig. \ref{figoxfordoutdoor} and Fig.\ref{figoxfordindoor} are examples of high-scoring detection on Oxford hand database for outdoor and indoor images, respectively. Obviously, our method works well for both outdoor and indoor images, for images with multiple and single hands. We give examples of false alarm detection in Fig.\ref{figoxfordfalse}, which indicates that skin areas(such as face, arm, foot) are more likely to be misunderstood as hand due to similar skin color, and some non-skin-like regions are also easy to be misclassified. We believe that we can make the hard negative more effective by running a skin area detection\cite{skinmodel} and intentionally add negative proposals from the skin area into the training data.

\subsubsection{Efficiency}
We compare the running time with the previous state-of-the-arts method\cite{mittal2011hand}, R-CNN\cite{girshick2014rich}, DP-DPM\cite{girshick2015deformable} in Table \ref{tabdetectiontime}. The time efficiency of our method is superior to that of the method in\cite{mittal2011hand}, and it is comparable to that of R-CNN and DP-DPM. The running time is about 8 seconds per image of $500\times 400$ pixels on a quad-core 2.9GHz PC with Nvidia Titan X, while previous method in \cite{mittal2011hand} takes about 55 seconds per image. Our method is more efficient due to the use of region proposal instead of sliding window, and derotating only once with estimated angle instead of brute force rotating in \cite{mittal2011hand}. We believe that our method can be more time efficient by leveraging more advanced region proposal method such as region proposal networks\cite{ren2015faster} and sharing convolutional feature maps of an image for all proposals by using pooling methods such as ROI pooling\cite{girshick2015fast}.
\begin{table}[t]
\caption{Average time (second/image) to detect hands. Comparison are made with state-of-the-arts DPM-based method\cite{mittal2011hand}, R-CNN\cite{girshick2014rich}, DP-DPM\cite{girshick2015deformable}, and our joint model. Our method is superior to \cite{mittal2011hand} in running time.} \centering
\begin{tabular}{c|cccc}
  \hline
  {} & DPM & R-CNN & DP-DPM & Joint model\\
  \hline
  running time &  55  & 9  & 2 & 8\\
    \hline
\end{tabular}
\label{tabdetectiontime}
\end{table}


\subsection{Model Analysis}
\subsubsection{Is The Model Well Optimized?}
In order to understand if the model is properly optimized with explicit rotation estimation, we train a detection network with the ground truth rotation. The precision-recall curve is shown in Fig. \ref{fig:precision_recall}(b). The average precision is 50.9\%, which can be considered as a performance upper bound under the current network topology. Again, it shows that aligning data to supervised orientation could great benefit the detection model. Also, our performance is only 2.6\% lower than this upper bound, which indicates our system is well optimized.

\subsubsection{Does Joint Training Help?}
Conceptually, it is beneficial to train a network by jointly optimizing over multiple related tasks. We investigate this issue here by comparing a model without jointly training to our joint model. To obtain a non-jointly optimized model, we still follow the divide and conquer fashion of parameter initialization, but allow the rotation network and the detection network to have shared first 3 layers for feature extraction. This results in 2\% drop on average precision(Refer to Fig. \ref{fig:precision_recall}(b)) and about 1\% drop on rotation estimation(Refer to Table \ref{tablerotation}). Overall, we demonstrate that joint training allows multiple tasks to share mutually useful information, and provides a compact and efficient network topology.


\subsection{Performance on EgoHands dataset}
\begin{figure*}[t]
\centering
\includegraphics[width=0.8\textwidth]{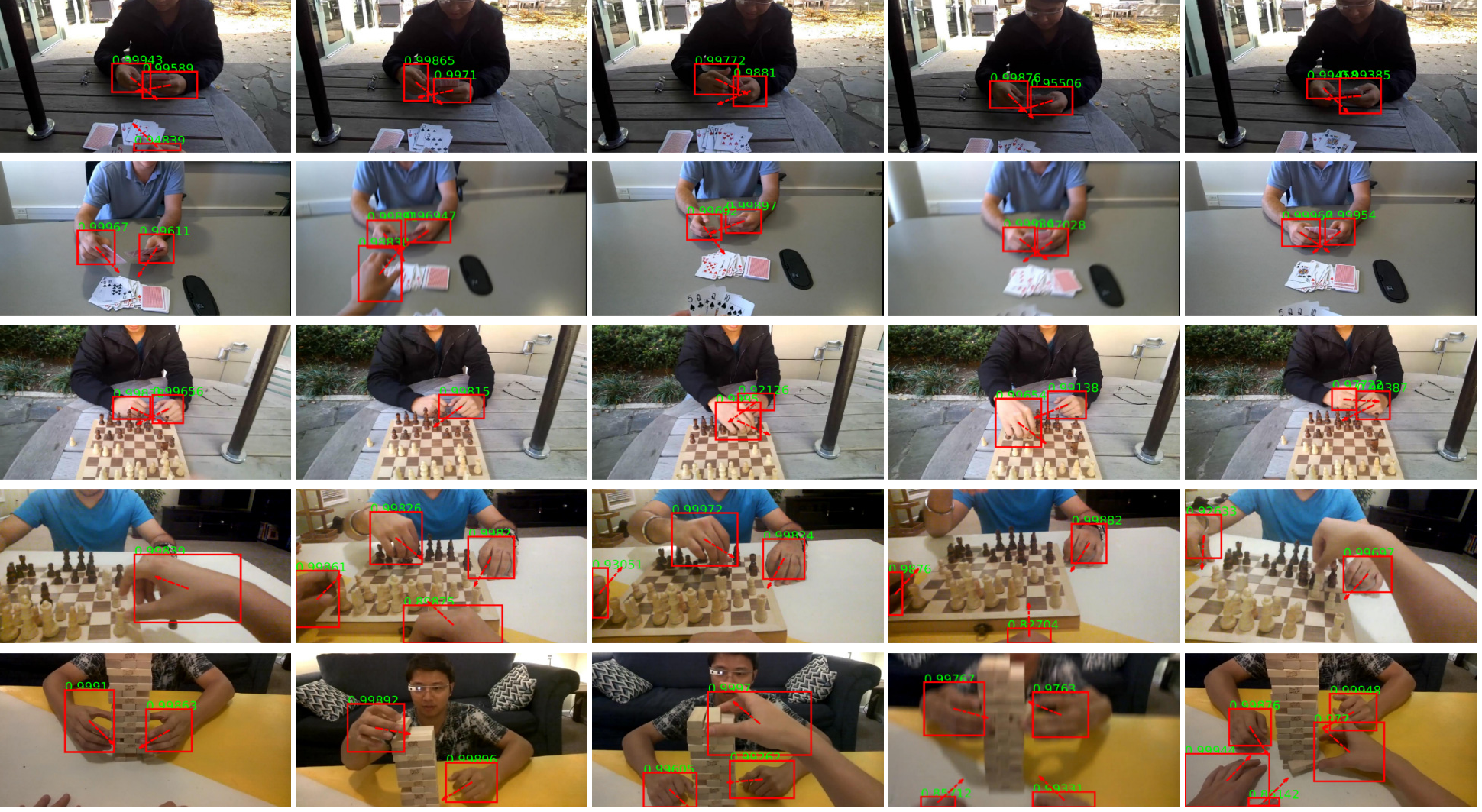}
\caption{Examples of high-scoring detection on Egohands database. The rotation estimation is illustrated with red arrows.}
\label{figegomoreresult}
\end{figure*}
\begin{figure}[htbp]
\centering
\includegraphics[width=0.5\textwidth]{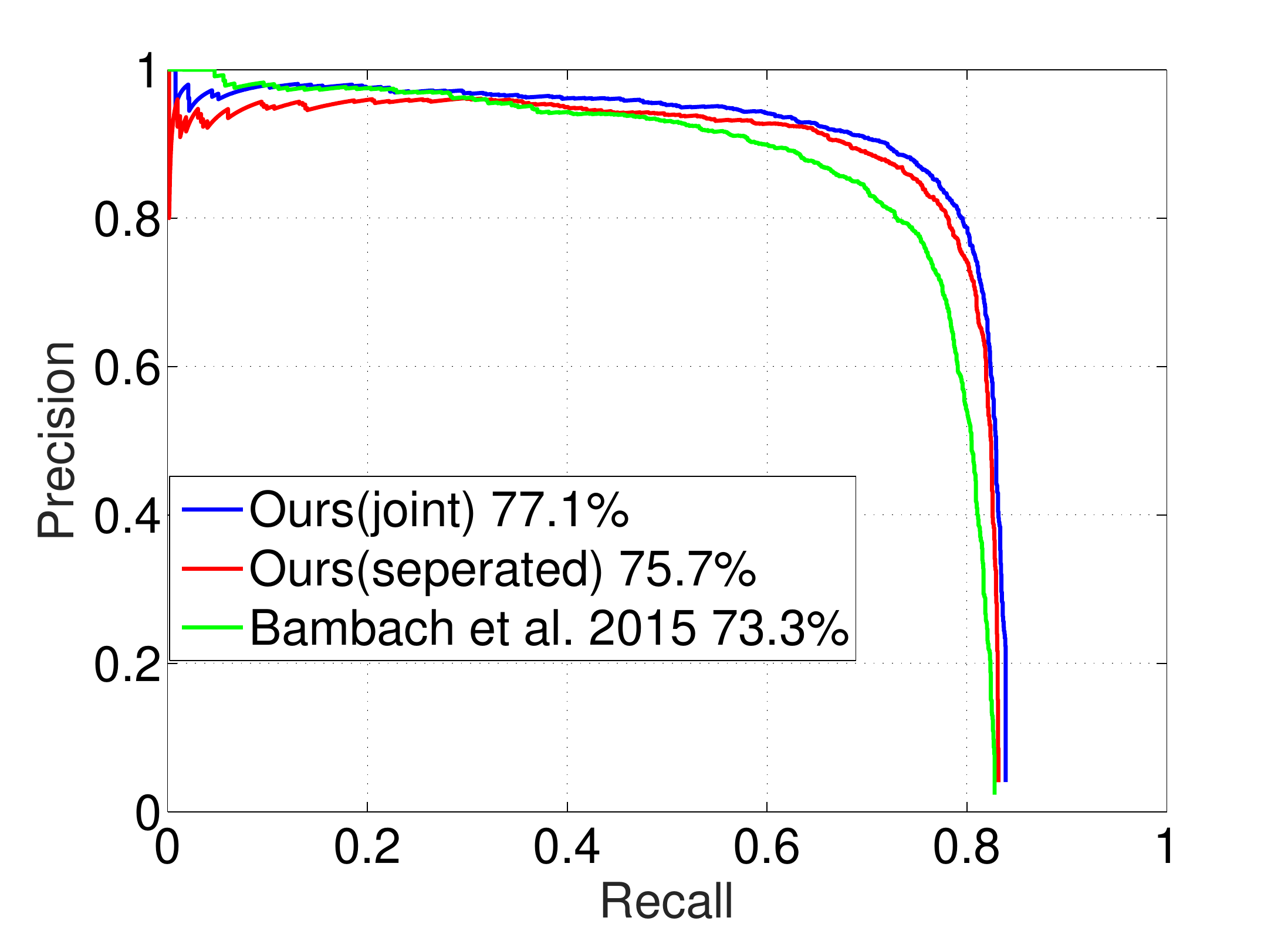}
\caption{Precision-Recall curve comparing the baseline and final results on EgoHands dataset\cite{egohands2015iccv}.}
\label{fig_ego_pccurve}
\end{figure}
In order to show the generalization power of our method, we test our pipeline on EgoHands dataset\cite{egohands2015iccv}. Fig.\ref{fig_ego_pccurve} shows precision-recall curve comparing the baseline and final results on EgoHands dataset, and the number after each algorithm is the average precision(AP). Our model(seperated) means that the shared 3 convolution layers are kept unchanged, and the rotation and detection networks are trained separately with shared network not tuned, and our model(joint) means that the network is end-to-end trained. Fig. \ref{figegomoreresult} shows examples of high-scoring detection on Egohands database. The rotation estimation performance on Egohands dataset are shown in Tab.\ref{tableegorotation}.

We compared our pipeline with the state-of-the-art detection algorithm in \cite{egohands2015iccv}. We implement the state-of-the-art hand detector on this dataset with the network prototxt and Caffemodel provided by \cite{egohands2015iccv}. For more rigorous evaluation, we compare detection performance of the method in  \cite{egohands2015iccv} and our method with the same region proposals, NMS and the other experiment setting. The average precision with our seperated model (AP:75.7\%) is higher than the results with baseline (AP:73.3\%)(Refer to Fig.\ref{fig_ego_pccurve}), which indicates that rotation information is helpful to improve the detection results. We then compare the rotation estimation and detection results with separated and joint models. We can see that the rotation estimation results with our joint model is slightly better than the results with only rotation model. Separated model results in 1.4\% drop on average precision than joint model. Therefore, we again demonstrate that joint training allows multiple tasks to share mutually useful information, and provides a compact and efficient network topology.
\begin{table}[t]
\caption{Rotation estimation performance on Egohands dataset.} \centering
\begin{tabular}{c|ccc}
  \hline
  {Method} & $\leq 10^\circ$ & $\leq 20^\circ$ & $\leq 30^\circ$ \\
  \hline
  Only rotation model &  48.63\% & 76.56\% & 87.26\%\\
  Joint model    &  49.01\% & 76.68\% & 87.09\%\\
  \hline
\end{tabular}
\label{tableegorotation}
\end{table}


\section{Conclusion}
Hand detection and pose estimation are important tasks for interaction applications. Previous works mostly solved the problem as separated tasks. In this paper, we explore the feasibility of joint hand detection and rotation estimation with CNN, which is based on our online derotation layer planted in the network. Our experimental results demonstrate that our method is capable of state-of-the-art hand detection on widely-used public benchmarks. The detection network can be extended to use hand context and more sophisticated rotation model.

\ifCLASSOPTIONcompsoc
  \section*{Acknowledgments}
\else
  \section*{Acknowledgment}
\fi
The authors would like to thank Dr. Peng Wang and Dr. Jiewei Cao from University of Queensland, and Dr. Sven Bambach from Indiana University Bloomington for helpful discussions.
The authors acknowledge the support of NVIDIA with the donation of the GPUs used for this research.

\bibliographystyle{IEEEtran}
\bibliography{IEEEabrv,refbib_yy}

\appendices
\section{Preliminary Analysis on ST-CNN}

We first show that ST-CNN has multiple comparative local optima under different transformation. Take affine transformation as an example, the point-wise transformation layer within ST-CNN is formulated as $x^s=\mathbf{A_\theta} x^t$, where $x^t$ is the target coordinates of the regular grid in the output feature map, $x^s$ is the source coordinates of the input feature map that define the sampling points, and $\mathbf{A}_\theta$ is the affine transformation matrix to optimize.


Suppose $\mathbf{A}_\theta$ after optimization aligns input feature maps into a certain pose. Denote $\mathbf{B}_\beta$ is an arbitrary 2D affine transformation, and obviously $\mathbf{B}_\beta\mathbf{A}_\theta$ can also align feature maps, but in different target poses. As a result, the output feature maps via $\mathbf{A}_\theta$ and $\mathbf{B}_\beta\mathbf{A}_\theta$ are not the same but both aligned. The detection networks trained with two sets of aligned features would have different network weights, but are very likely to have similar detection performance. Therefore, the loss function could reach comparative local minima with either $\mathbf{A}_\theta$ or $\mathbf{B}_\beta\mathbf{A}_\theta$.

We now know that many combinations of transformation parameters and detection weights could result in similar detection performance, i.e. ambiguous rotation estimation and many local minima. The transformation space is typically huge and would require much more data and time to converge. We adopt a supervised approach to get the rotation parameters. Our network will not wonder back and forth between ambiguous transformations, but insists on moving towards the desired pose.

\begin{figure}[t]
\centering
\includegraphics[width=0.5\textwidth]{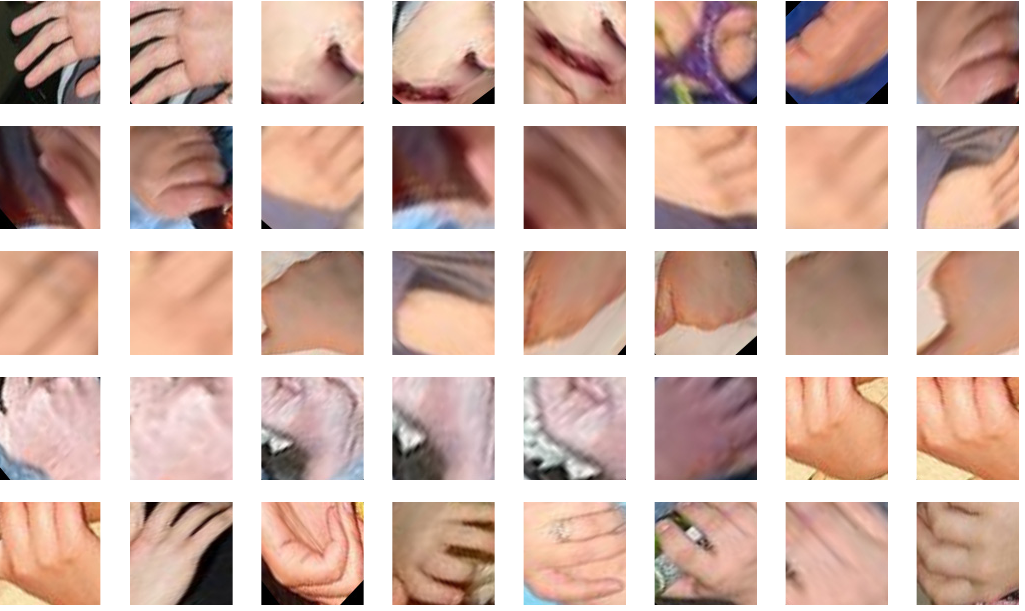}
\caption{Hand proposals transformed with affine transformation obtained by ST-CNN.}
\label{figstcnn}
\end{figure}

We conduct hand detection experiment with ST-CNN. We add a spatial transformation layer after the input data layer of an AlexNet. Fig.\ref{figstcnn} shows hand proposals transformed with affine transformation via ST-CNN. It shows that the hand proposals are not well aligned. In fact, from the result we can see that the ST-CNN fails to learn the transformation that align input proposals, but retreat back to a trivial translation that only captures the major part of the object, i.e. palm region in our case, which is a bad local optima. While the transformed proposals can be still used for the detection network followed, key hand context information is missing (The importance of context for hand and generic object detection is elaborated in \cite{mittal2011hand}\cite{divvala2009empirical}). Therefore, the detection performance with ST-CNN could be poor(Please refer to Fig.~7(a). The performance of ST-CNN is even worse than R-CNN in hand detection task). In summary, for hand detection task, ST-CNN is prone to learn ambiguous transformation, resulting images often miss key context information, which may not be the ideal model for hand detection.

\ifCLASSOPTIONcaptionsoff
  \newpage
\fi

\end{document}